\documentclass[letterpaper, 10 pt, conference]{ieeeconf}  
\IEEEoverridecommandlockouts                              
\usepackage[english]{babel}
\usepackage[T1]{fontenc}

\let\labelindent\relax
\usepackage{enumitem}
\usepackage{balance}
\usepackage[font={small}]{caption}
\usepackage{subcaption}
\usepackage{array}
\usepackage{textcomp}
\usepackage{mathtools, nccmath}
\usepackage{graphicx}
\usepackage{amsfonts}
\usepackage{amsmath}
\usepackage{amssymb}
\usepackage{algorithm}
\usepackage{algorithmic}
\usepackage{hyperref}
\usepackage{tikz}
\usepackage{arydshln}
\usepackage{multirow}
\usepackage{bm}
\usepackage{epstopdf}
\usepackage{cite}
\usepackage{pifont}
\usetikzlibrary{positioning}

\title{\LARGE \bf Reinforcement Learning for Robust Parameterized \\ Locomotion Control of Bipedal Robots}

\author{Zhongyu Li, Xuxin Cheng, Xue Bin Peng, Pieter Abbeel, Sergey Levine, Glen Berseth, Koushil Sreenath
\thanks{All authors are with the University of California, Berkeley, CA, USA.
{\tt\footnotesize \{zhongyu\_li, chengxuxin, xbpeng, gberseth\}@berkeley.edu, pabbeel@cs.berkeley.edu, svlevine@eecs.berkeley.edu, koushils@berkeley.edu}}
\thanks{This work is supported in part by National Science Foundation Grants CMMI-1944722 and IIS-1651843, the Office of Naval Research, NSERC Postgraduate Scholarship, a Berkeley Fellowship for Graduate Study, and BAIR.}}

\begin{document}
\maketitle

\begin{abstract}
Developing robust walking controllers for bipedal robots is a challenging endeavor.
Traditional model-based locomotion controllers require simplifying assumptions and careful modelling; any small errors can result in unstable control.
To address these challenges for bipedal locomotion, we present a model-free reinforcement learning framework for training robust locomotion policies in simulation, which can then be transferred to a real bipedal Cassie robot.
To facilitate sim-to-real transfer, domain randomization is used to encourage the policies to learn behaviors that are robust across variations in system dynamics.
The learned policies enable Cassie to perform a set of diverse and dynamic behaviors, while also being more robust than traditional controllers and prior learning-based methods that use residual control.
We demonstrate this on versatile walking behaviors such as tracking a target walking velocity, walking height, and turning yaw. (Video\footnote{ Video: \urlstyle{same}\url{https://youtu.be/goxCjGPQH7U}})
\end{abstract}

\section{Introduction}

\label{sec:introduction}

Many environments, particularly those designed for humans, are more accessible by legged systems.
However, bipedal robot locomotion involves several control design challenges due to high degrees-of-freedom (DoFs), hybrid nonlinear dynamics, and persistent but hard-to-model ground impacts. 
Classical model-based methods~\cite{vukobratovic2004zero,koolen2012capturability,GrChAmSi2010} for stabilizing and controlling bipedal systems tend to require careful modeling and usually lack the ability to adapt to changes in the environment.
Recent deep reinforcement learning~(RL) based methods are promising solutions to these issues as RL is able to leverage the full-order dynamics of the system to produce more agile behaviors.

Our approach to bipedal locomotion utilizes RL to train robust policies to imitate gaits from a gait library, using randomized simulated training to acquire controllers that can successfully control a person-sized bipedal robot Cassie in the real world.
Recent RL methods on Cassie~\cite{xie2018feedback,xie2020learning} train policies to specify corrections to reference motions recorded from a model-based walking controller. 
While this type of residual control~\cite{johannink2019residual} can produce stable walking behaviors, it often requires a pre-existing controller, and the resulting behaviors tend to be limited to stay close to the original reference motions. 
To overcome this limitation, our training system uses a gait library of diverse parameterized motions based on Hybrid Zero Dynamics~(HZD)~\cite{GrChAmSi2010}, which increases the diversity of behaviors that the robot can learn. 
This increase in diversity serves two purposes. 
First, it enables the online control of the robot using a low dimensional parameterization of different gaits. Second, it improves robustness by enabling the robot to learn a larger variety of potential motions.

\begin{figure}[t]
    \centering
    \begin{subfigure}{.49\linewidth}
    \includegraphics[width=1\linewidth]{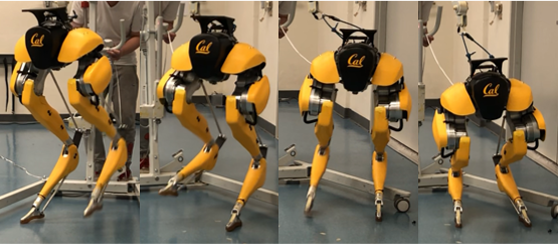}
    \caption{Lower Walking Height}
    \label{subfig:lower_height}    
    \end{subfigure}
    \begin{subfigure}{.49\linewidth}
    \includegraphics[width=1\linewidth]{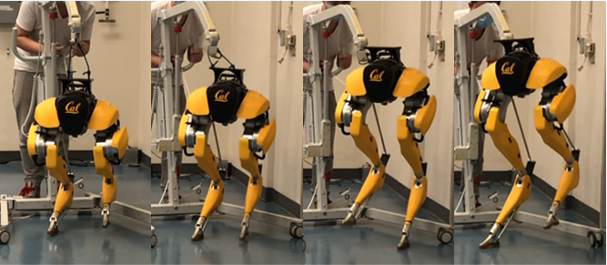}
    \caption{Recover to Normal Height}
    \label{subfig:enlarge_height}    
    \end{subfigure}
    \begin{subfigure}{.49\linewidth}
    \includegraphics[width=1\linewidth]{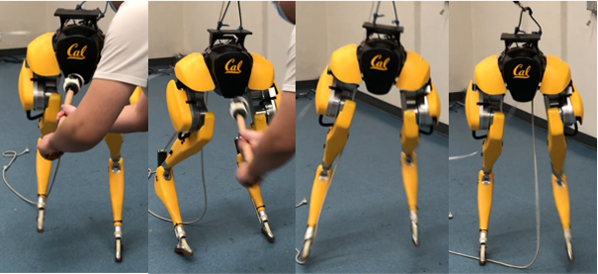}
    \caption{Push Recovery (Front)}
    \label{subfig:recover_front}    
    \end{subfigure}
    \begin{subfigure}{.49\linewidth}
    \includegraphics[width=1\linewidth]{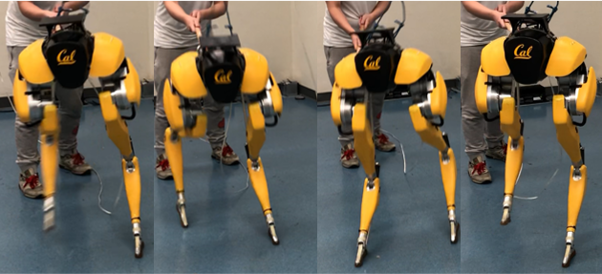}
    \caption{Push Recovery (Back)}
    \label{subfig:recover_back}    
    \end{subfigure}
    \caption{Our system leverages reinforcement learning to train robust parameterized locomotion controllers for a bipedal Cassie robot. The controller can vary parameters such as walking velocity and height, while also being robust to significant external perturbations.}
    \label{fig:main}
    \vspace{-0.5cm}
\end{figure}

Due to the size and instability of real bipedal robots, it is particularly dangerous to perform RL directly on the physical system.
We instead leverage techniques from sim-to-real transfer to train policies in a simulated environment, which are then evaluated in another higher fidelity simulator, before finally being deployed on the real robot. 
We found that additional dynamics randomization techniques are necessary to produce robust controllers, and these enable our system to learn robust parameterized controllers, which are then evaluated on Cassie with a collection of different motions and challenging scenarios, as shown in Fig.~\ref{fig:main}.

The primary contribution of this work is the development of a reinforcement learning based controller as shown in Fig.~\ref{fig:controllers} that results in a more diverse and robust walking control on the Cassie robot.
More specifically, we develop an end-to-end versatile walking policy that combines a HZD-based gait library with deep reinforcement learning to enable a $3D$ bipedal robot Cassie to walk while following commands for frontal and lateral walking speeds, walking height, and turning yaw rate.
The proposed learning-based walking policy notably expands the feasible command set and safe set over prior model-based controllers, and improves stability during gait transitions compared to a HZD-based baseline walking controller. 
The learned policies are robust to modelling error, perturbations, and environmental changes.
This robustness emerges from our training strategy, which trains a policy to imitate a collection of diverse gaits, and also incorporates domain randomization.
The learned policy can be directly transferred to other simulators, such as a more accurate simulator on SimMechanics, as well as to a real robot. 
Using this proposed policy, Cassie is not only able to reliably track given commands in indoor and outdoor environments, but to stay robust to unmodelable malfunctioning motors, changes of ground friction, carrying unknown loads, and demonstrating agile recoveries from random perturbations, as shown in the video.

\subsection{Related Work}

Traditional approaches for locomotion of bipedal walking robots are typically based on \textit{notions of gait stability},
such as the ZMP criterion~\cite{vukobratovic2004zero} and capturability~\cite{koolen2012capturability}, simplified models~\cite{kuindersma2014efficiently,pratt2012capturability,xiong2018coupling}, and constrained optimization methods~\cite{koolen2013summary,feng20133d,dai2014whole}. 
These methods have been shown to be effective for controlling various humanoid robots with flat feet, but the resulting motion tends to be slow and conservative. 
Hybrid Zero Dynamics~(HZD)~\cite{GrChAmSi2010,hereid2018rapid,da20162d, RSS2017_DiscreteTerrain_Walking, gong2019feedback, li2020animated} is another control technique for generating stable periodic walking gaits based on input-output linearization.
Our work, which is based on reinforcement learning, is not constrained by the requirement of a precise model and stabilization to a periodic orbit, as is the case for HZD, which enables our method to produce more diverse behaviors.

\paragraph{RL-based Control for Legged Robots}
Reinforcement learning for legged locomotion has shown promising results in acquiring locomotion skills in simulation~\cite{coros2011locomotion,peng2017deeploco,peng2018deepmimic} and in the real world~\cite{kohl2004policy,hwangbo2019learning,peng2020learning}. 
Data-driven methods provide a general framework that enables legged robots to perform a rich variety of behaviors by introducing reference motion terms into the learning process~\cite{ratliff2007imitation,peng2018deepmimic,peng2020learning}.
However, most previous RL-based work are deployed on either multi-legged systems~\cite{li2020learning,peng2020learning,lee2020learning} or on low-dimensional bipedal robots~\cite{morimoto2007learning,yu2019sim}, where learned motions are typically quasi-static.

More recently, RL has been applied to learn agile walking skills for Cassie. 
Model-based RL in~\cite{castillo2020velocity, castillo2020hybrid} attains a velocity regulating adaptive walking controller on Cassie in simulation. 
In~\cite{xie2018feedback,xie2020learning}, reference motions combined with model-free residual learning~\cite{johannink2019residual} are used to learn walking policies that are able to reliably track given planar velocity commands on Cassie in the real world. 
Residual control structure used in the policy can speed up training, but the resulting policy can only apply limited corrections to the underlying reference trajectory. 
Moreover, a model-based walking controller on Cassie is still needed to provide reference motions recorded from control outputs.
This limits the accessibility to the reference motions and therefore reduces the diversity of learned behaviors.
In addition, most of the previous learning-based walking policies on Cassie do not show significant improvement over traditional model-based controllers. 
Also, they lack the ability to change the walking height and turning yaw, which increases the complexity of controller design but enables the robot to travel in narrow environments.
In our work, we show a clear improvement over model-based methods by examining the tracking performance and robustness over a range of gait parameters.

\paragraph{Simulation to Real World Transfer}
Sim-to-real transfer is an attractive approach for developing policies, which takes advantage of fast simulations as a safe and inexpensive source of data. 
Model-based methods require careful system identification to bridge the reality gap~\cite{gong2019feedback,li2020animated}. 
Randomizing the system properties in the source domain in order to cover the uncertainty in the target domain has allowed for solutions that use low-fidelity simulations for learning-based methods~\cite{sadeghi2016cad2rl,tobin2017domain,peng2018sim,yu2019sim,yu2020learning,peng2020learning,siekmann2020learning}. 
In this paper, we adopt domain randomization to overcome the sim-to-real gap, without the need for any additional training on the robot.

\section{Parameterized Control of Cassie}
\label{sec:background}
We now present the Cassie bipedal robot, which is the platform for our experiments and introduce a HZD-based gait library of versatile walking behaviors on Cassie. 

\subsection{Cassie Robot Model}


Cassie is a person-sized, dynamic, underactuated bipedal robot with $20$ DoFs, as shown in Fig.~\ref{fig:main} and explained in~\cite[Sec. II]{li2020animated}. There are 10 actuated rotational joints $q_m = [q_{1,2,3,4,7}^{L/R}]^T$, which include abduction, rotation, hip pitch, knee, and toe motors. There are also four passive joints $q_{5,6}^{L/R}$ that correspond to the shin and tarsus joints. Its floating base pelvis $q_p = [q_{x}, q_{y}, q_{z}, q_{\psi}, q_{\theta}, q_{\phi}]^T$ has 3 transitional DoFs~(sagittal, lateral, vertical) $q_{x,y,z}$ and 3 rotational DoFs~(roll, pitch, yaw) $q_{\psi,\theta,\phi}$, and is defined as the robot's local reference frame. The full robot state $q \in \mathbb{R}^{20}$ consists of the state of each joint. 
Next, we define the observable state $q^o \in \mathbb{R}^{17}$, which is similar to $q$ but excludes the pelvis translational position $q_{x,y,z}$, that can not be reliably measured in the real world without external instrumentation.

\begin{figure}[t]
    \centering
    \begin{subfigure}{\linewidth}
    \includegraphics[width=1\linewidth]{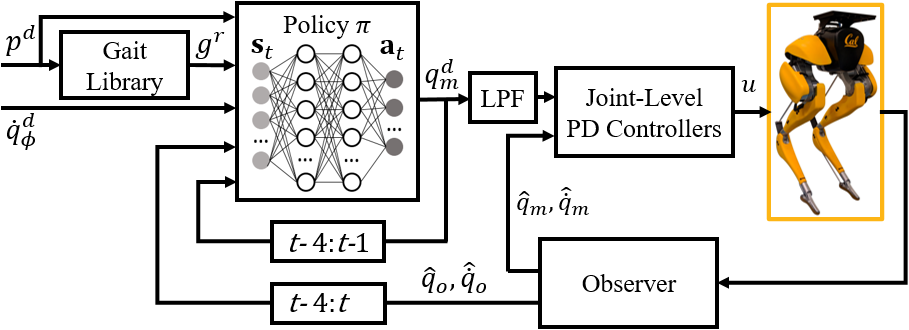}
    \label{subfig:learned_controller}    
    \end{subfigure}
    \caption{Proposed learning-based walking controller. The inputs of the policy consists of desired gait parameter $p^d$, desired turning yaw velocity $\dot{q}^d_{\phi}$, a reference gait $g^r$ decoded by desired gait parameter $p^d$, observed robot states $\hat{q}_o$, $\hat{\dot{q}}_o$ from time step $t$-4 to $t$, and past policy outputs which are desired motor positions $q^d_m$ spanning from $t$-4 to $t$-1. Current $q^d_m$ is sent to joint-level controllers after passing through a Low Pass Filter~(LPF).}
    \label{fig:controllers} 
    \vspace{-0.5cm}
\end{figure}

\subsection{Gait Library and Parameterized Control}
\label{subsec:gaitlibrary}
To create a controller that can be directed online to perform and transition between different motions, we parameterize the input to the system using a \emph{gait parameter} $p$ that determines the desired \textit{gait}.
A \textit{gait} $g$ is a set of periodic joint trajectories that encode a locomotion behavior~\cite{haynes2006gaits}. In this work, we use $5^{th}$ order B\'ezier curves $\alpha$ to represent smooth profiles for the $10$ actuated joints. The B\'ezier curves are normalized by $1$ step period $\overline{t}$. The gaits designed in this paper consist of $2$ steps, referred to as \emph{right stance} and \emph{left stance}, and transitions between the steps are triggered by a foot \emph{impact} on the ground. 
The gait parameters chosen in this work are forward velocity $\dot{q}_x$, lateral velocity $\dot{q}_y$ and walking height $q_z$, \textit{i.e.}, $p = [\dot{q}_x~\dot{q}_y~q_z]^T \in \mathbb{R}^3$.
A \textit{gait library} $\mathcal{G} = \{g_i(\alpha,\overline{t})\}$ is constructed by indexing the $i^{th}$ gait $g_i$ with its gait parameter $p_i$. The optimization program for constructing the HZD-based gait library is formulated in CFROST~\cite{hereid2018rapid} and the resulting gaits are described in Tab.~\ref{tab:gaitlibrary}~\cite{li2020animated}. 
The gait libray is later combined with an online regulator to implement a parameterized walking controller in~\cite[Sec. IV]{li2020animated}.
\section{Learning Walking Control and Sim-to-Real}
\label{sec:method}


\begin{table}[!]
\centering
\caption{Gait Library}
\label{tab:gaitlibrary}
\begin{tabular}{ccc}
\hline
 & $\dot{q}_x$  & $\dot{q}_y$ \\ \hline
$\mathcal{G}$ & ${[}-1,-0.8,\dots,0.8,1.0{]}$ & ${[}-0.3,-0.24,\dots,0.24,0.3{]}$ \\
\hline
 & $q_z$ & number of gaits \\ \hline
$\mathcal{G}$ & ${[}0.65,0.685,\dots,0.965,1.0{]}$ & $11\times 11\times 11=1331$  
\\
\hline
\end{tabular}
\vspace{-0.5cm}
\end{table}

Having an optimized gait library is not enough to control bipedal robots without online feedback, we will next combine the pre-computed HZD-based gait library with reinforcement learning to develop a versatile locomotion policy $\pi$ for the Cassie.
In a RL framework, an agent (\textit{e.g.} Cassie), learns through trial and error by interacting with the environment.
At each discrete time step $t$, the policy $\pi$ (shown in Fig.~\ref{fig:controllers}) observes a state $\mathbf{s}_t$ and a goal $\mathbf{g}_t$, and outputs an action distribution $\pi(\mathbf{a}_t|\mathbf{s}_t,\mathbf{g}_t)$. 
The agent then samples an action $\mathbf{a}_t$ from the distribution and executes the action in the environment, which results in a transition to a new state $\mathbf{s}_{t+1}$ and goal $\mathbf{g}_{t+1}$, as well as a reward $r_t$ for that transition.

\subsection{Cassie Simulation Environment}
\label{subsec:action_state_goal}
We developed a simulation environment for reinforcement learning on Cassie, which is based on an open source MuJoCo simulator\cite{todorov2012mujoco,cassiesim}.
This subsection introduces the design of the simulation environment which the reinforcement learning agent interacts with.
\subsubsection{Action Space}
\label{subsubsec: action}
The action $\mathbf{a}_t = q_m^d$ specifies target positions for the 10 motors on Cassie. 
In order to obtain a smoother motion, the target positions are first passed through a low-pass filter~\cite{peng2020learning}, as shown in Fig.~\ref{fig:controllers}, before being applied to the motors. 
A joint-level PD controller generates torque $u$ for each motor on Cassie based on the filtered targets.

\subsubsection{State Space}
The state $\mathbf{s}_t = (\mathbf{q}^o_{t-4:t}, \mathbf{a}_{t-4:t-1})$ at time $t$ consists of two components. The first component consists of the observable robot states $\mathbf{q}^o=[q^o,\dot{q}^o]$ at the current time step $t$ and the past 4 time steps.
The second component consists of the actions $\mathbf{a}$ from past 4 time steps. 
This history of past observations and actions provides the policy with more information to infer the system dynamics. 

\subsubsection{Goal}
\label{subsubsec:goal}
To train a policy to produce a desired reference motion, target frames from the reference motion are provided to the policy as input via a time-dependent goal $\mathbf{g}_t$.
The user command $c$ is used to operate the robot online and it is defined as $c=[p^d~\dot{q}^d_{\phi}]=[\dot{q}^d_x~\dot{q}^d_y~q^d_z, \dot{q}^d_{\phi}]$
which includes desired gait parameters $p^d$ and desired turning yaw velocity $\dot{q}^d_{\phi}$.
Given a desired gait parameter $p^d$, a reference gait $g^r$ is constructed by interpolating the parameterized gait library with respect to $p^d$ as explained in Sec.~\ref{subsec:gaitlibrary}. 
The goal is then specified by $\mathbf{g}_t = (c(t), g^r(t), g^r(t+1), g^r(t+4), g^r(t+7))$, which includes 1) the current user commands, and 2) reference motor positions $q^m_r$ and velocity $\dot{q}_m^r$ for current and sampled future time steps.

\subsection{Reward Function}
The reward function is designed to encourage the agent to satisfy the given command while reproducing the corresponding reference motion from the gait library on the dynamic robot system. The reward at each time step $t$ is given by:

\begin{align}
    r_t &= \omega r_t^o \\
    r_t^o &= [r_t^m, r_t^{p}, r_t^{\dot{p}}, r_t^{r},  r_t^{\dot{r}}, r_t^u, r_t^f]^T\\
    \omega &= [0.3, 0.24, 0.15, 0.13, 0.06, 0.06, 0.06]
\end{align}
The motor reward $r_t^m$ encourages the policy to minimize the discrepancies between the actual motor positions $\hat{q}_m$ and the reference motion $q_m^r$ and is formulated as:
\begin{equation}\label{eq:motor_reward}
    r_t^m = \exp[-\rho_1||q_m^r - \hat{q}_m||^2_2] .
\end{equation}
where $\rho_i$ is a scaling factor for the $i^{th}$ reward term. 
The reward terms $r_t^{p}$, $r_t^{\dot{p}}$ and $r_t^{\dot{r}}$ follow the same formulation as \eqref{eq:motor_reward} and encourage the agent to track reference pelvis translational position, translational velocity and rotational velocity in robot local frame, respectively.
The pelvis rotation reward $r_t^{r}$ leads Cassie to reduce the difference between the reference rotation $q_r^r$ and the actual one $\hat{q}_r$, and it is formulated by $r_t^{r} = \exp[-\rho_4||q_r^r \ominus \hat{q}_r||^2_2]$ where $\ominus$ denotes the geodesic distance between two rotation angles. 
The torque reward $r_t^u = \exp[-\rho_6||u||^2_2]$ encourages the robot to reduce energy consumption. 
Lastly, the ground reaction force reward $r_t^f = \exp[-\rho_7||\hat{q}_f||^2_2]$ helps to minimize the vertical contact forces $\hat{q}_f$. 
The weights of each reward term in $\omega_{i}$ are specified manually.

The desired roll and pitch velocity are always set to $0$ to stabilize the pelvis, while the desired yaw velocity is specified by the user command $c(t)$.
Furthermore, since no desired position terms, \textit{e.g.}, pelvis translational and rotational positions, are explicitly given, they are computed by integrating the corresponding desired velocity.

Note that the reference motion from the gait library does not encode turning yaw information. 
Therefore, including non-zero desired turning yaw in the reward can encourage the agent to develop walking behaviors that are not provided by the reference motions. 

\subsection{Domain Randomization}\label{subsec:domain_randomization}
In order to improve the robustness of the policy and bridge the gap between the simulation and the real world, the dynamics of the environment is randomized during training in simulation. The randomization regiment is designed to address three major sources of uncertainty in the environment: 1) modelling error of the robot and the environment, 2) sensor noise, and 3) communication delay between the policy and the joint-level controller. These dynamics properties are parameterized as $\bm{\mu}$, whose values are varied between the ranges specified in Tab.~\ref{tab:randomization}. 

\begin{table}[t]
\centering
\caption{Dynamics Properties and Sample Range.}
\label{tab:randomization}
\begin{tabular}{ccc}
\hline
\textbf{Parameter}         & \textbf{Range}            & \textbf{Unit}      \\ \hline
Link Mass                  & {[}0.75,1.15{]} $\times$ default & kg        \\
Link Mass Center           & {[}0.75,1.15{]} $\times$ default & m         \\
Joint Damping              & {[}0.75,1.15{]} $\times$ default & Nms/rad   \\
Ground Friction Ratio      & {[}0.5, 3.0{]}            & 1                \\
Motor Rotation Noise       & {[}-0.1, 0.1{]}           & rad             \\
Motor Angle Velocity Noise & {[}-0.1, 0.1{]}           & rad/s            \\
Accelerometer  Noise       & {[}-0.4, 0.4{]}           & m/s$^\text{2}$            \\
Gyro Rotation Noise        & {[}-0.1, 0.1{]}           & rad              \\
Gyro Angle Velocity Noise  & {[}-0.1, 0.1{]}           & rad/s           \\
Communication Delay        & {[}0, 0.03{]}             & sec             \\ \hline
\end{tabular}
\vspace{-0.5cm}
\end{table}

\subsection{Learning Model}
The objective of reinforcement learning is to maximize the total expected reward over trajectories $\tau \sim p_\theta(\tau)$
 \begin{equation}
     J(\theta)=\mathbb{E}_{\tau \sim p_\theta(\tau)}\left[\sum_{t=0}^{T} \gamma^{t} r_t\right]
    \label{equa:obj}
 \end{equation}
where $p_\theta(\tau)$ is the distribution of trajectories $\tau = \{\mathbf{s}_0, \mathbf{a}_0, r_0, \ldots, \mathbf{s}_T, \mathbf{a}_T, r_T\}$ subject to policy $\pi_\theta(\mathbf{a}_t| \mathbf{s}_t, \mathbf{g}_t)$, 
$\theta$ is the parameters of the network policy, $\gamma$ is the discount factor, and $T$ is the horizon of each episode.
We use Proximal Policy Optimization (PPO)\cite{schulman2017proximal} to train the policy in simulation with networks with 2 hidden layers of $512$ $tanh$ units for both the policy and value function. 
 
For the policy network, the input contains observed state $\mathbf{s}^o_t$ and goal $\mathbf{g}_t$, as formulated in Sec.~\ref{subsec:action_state_goal}. 
In $\mathbf{s}^o_t$, the observed robot state $\mathbf{q}^o_t$ is with added noise and delay introduced in Sec.~\ref{subsec:domain_randomization}. 
The policy network uses $tanh$ as the activation function for the last layer.
The output of the policy network is a $10$ dimensional vector represented
by a Gaussian action distribution $\mathcal{N}(\mathbf{m}_\pi(\mathbf{a}| \mathbf{s}), \mathbf{\Sigma}_\pi)$ with a learned mean $\mathbf{m}_\pi(\mathbf{a}| \mathbf{s})$ and fixed standard deviation $\mathbf{\Sigma}_\pi = 0.1I$. 
The action $\mathbf{a}$ (\textit{i.e.}, desired motor positions) is sampled from this output distribution.

The value network outputs a scalar value $V(\mathbf{s}_t, \mathbf{g}_t)$ representing the expected return of the policy given state $\mathbf{s}_t$ and goal $\mathbf{g}_t$.
The value network is provided access to the ground truth state $\mathbf{s}^{gt}$,
Moreover, the randomized dynamic parameters $\bm{\mu}$ described in Sec.~\ref{subsec:domain_randomization} are also provided as inputs to the value function.

\subsection{Training Setup}
The policy operates at $30$~Hz, while the joint-level PD controller illustrated in~Fig.~\ref{fig:controllers} runs at $2000$~Hz.
The maximum number of time steps for each episode is designated to be $T$=$2500$, corresponding to approximately $83$~s. 
In each episode, a new command $c(t)=[\dot{q}^d_x~\dot{q}^d_y~q^d_z, \dot{q}^d_{\phi}]$ is uniformly sampled every $8$~s, and remains unchanged during the $8$~s window. The command range is from $[-2,-0.8,0.65,-\pi/6]^T$ to $[2,0.8,1,\pi/6]^T$. The first command in each episode is always set to a random walking forward velocity with $0$ yaw velocity at a normal height above $0.9$~m. 
Note that the range of training commands is larger compared to the gait parameter provided in Tab.~\ref{tab:gaitlibrary}. 
In this way, the agent is able to learn to follow a command that is out-of-range of gait parameters and thus learns behaviors beyond what the gait library can provide.
An episode ends when the maximum number of time steps is reached, or early termination conditions have been triggered. 
Early termination is triggered if the height of the pelvis drops below $0.55$~m, and if the tarsus joints $q^{L/R}_5$ hit the ground. 

Dynamics randomization presented in Sec.~\ref{subsec:domain_randomization} is introduced gradually over the course of training through a curriculum. The curriculum helps to prevent the policy from adopting excessively conservative sub-optimal behaviors. For example, if training starts with the full range of randomizations detailed in Table~\ref{tab:randomization}, the policy is prone to adopting strategies that prevent the robot from falling by simply standing in-place. In a highly dynamic environment, standing can be more stable than walking, and is therefore easier to learn, but is nonetheless sub-optimal.
Therefore, over the course of the first $2000$ training iterations, the upper and lower boundaries of the randomized dynamics parameters are linearly annealed from fixed default values to the maximum ranges specified in Tab.~\ref{tab:randomization}.
\section{Experiments}
\label{sec:results}
The walking policy is trained with a MuJoCo simulation of the Cassie robot~\cite{todorov2012mujoco}. 
The performance of the learned policy is evaluated in three domains: MuJoCo, MATLAB SimMechanics, and on Cassie.
Performance is first evaluated in the MuJoCo simulator, which is the domain used for training. 
Later, SimMechanics provides a safe and high-fidelity simulated environment that closely replicates the physical system to extensively test the learned policy.
However, the high-fidelity simulation is slower than real-time by an order of magnitude, so it is primarily used for testing.
Finally, the policy is deployed and validated on the real Cassie robot. 

\subsection{Learning Performance}
\begin{figure}[t]
\begin{subfigure}{0.45\columnwidth}
    \centering
     \includegraphics[width=\columnwidth]{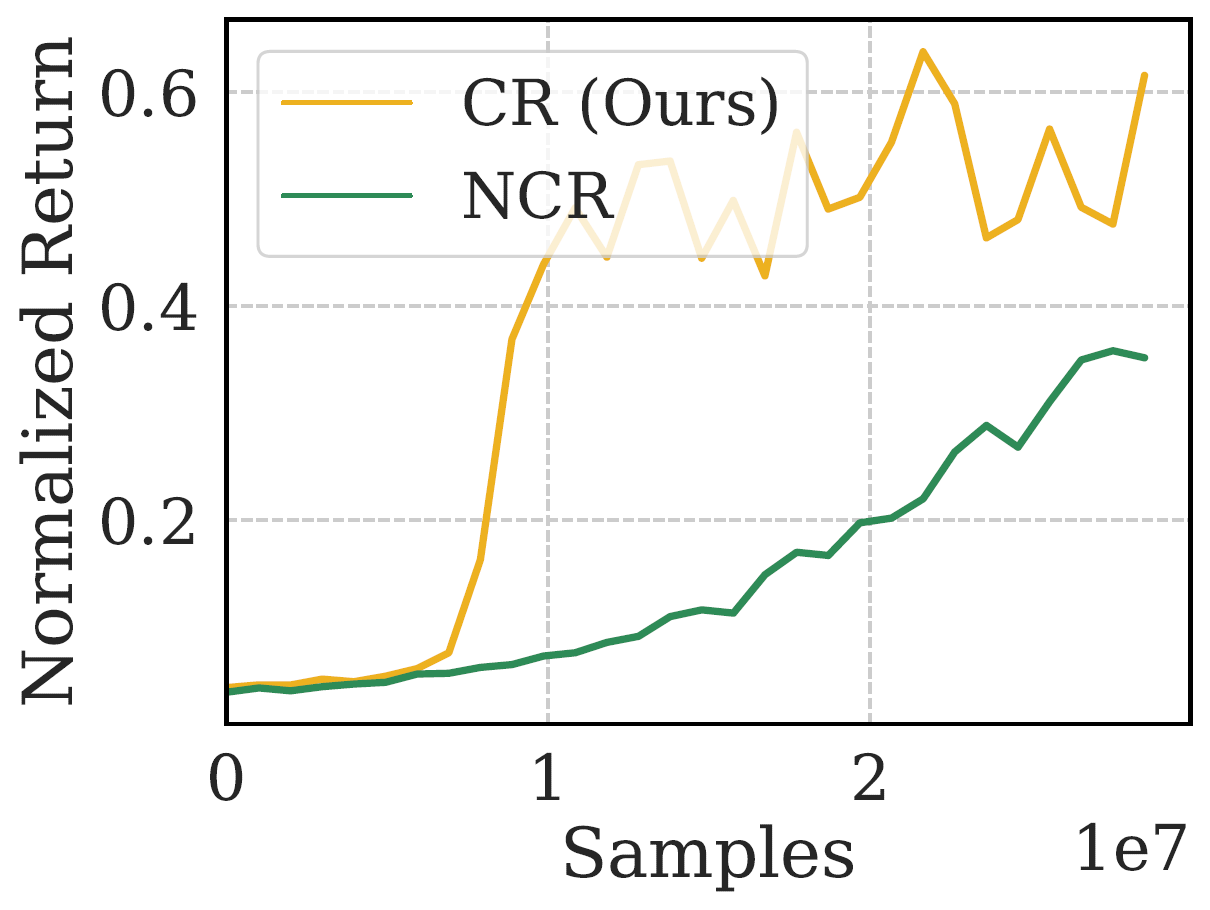}
     \caption{}
     \label{fig:reward_curr}
\end{subfigure}
\begin{subfigure}{0.45\columnwidth}
    \centering
     \includegraphics[width=\columnwidth]{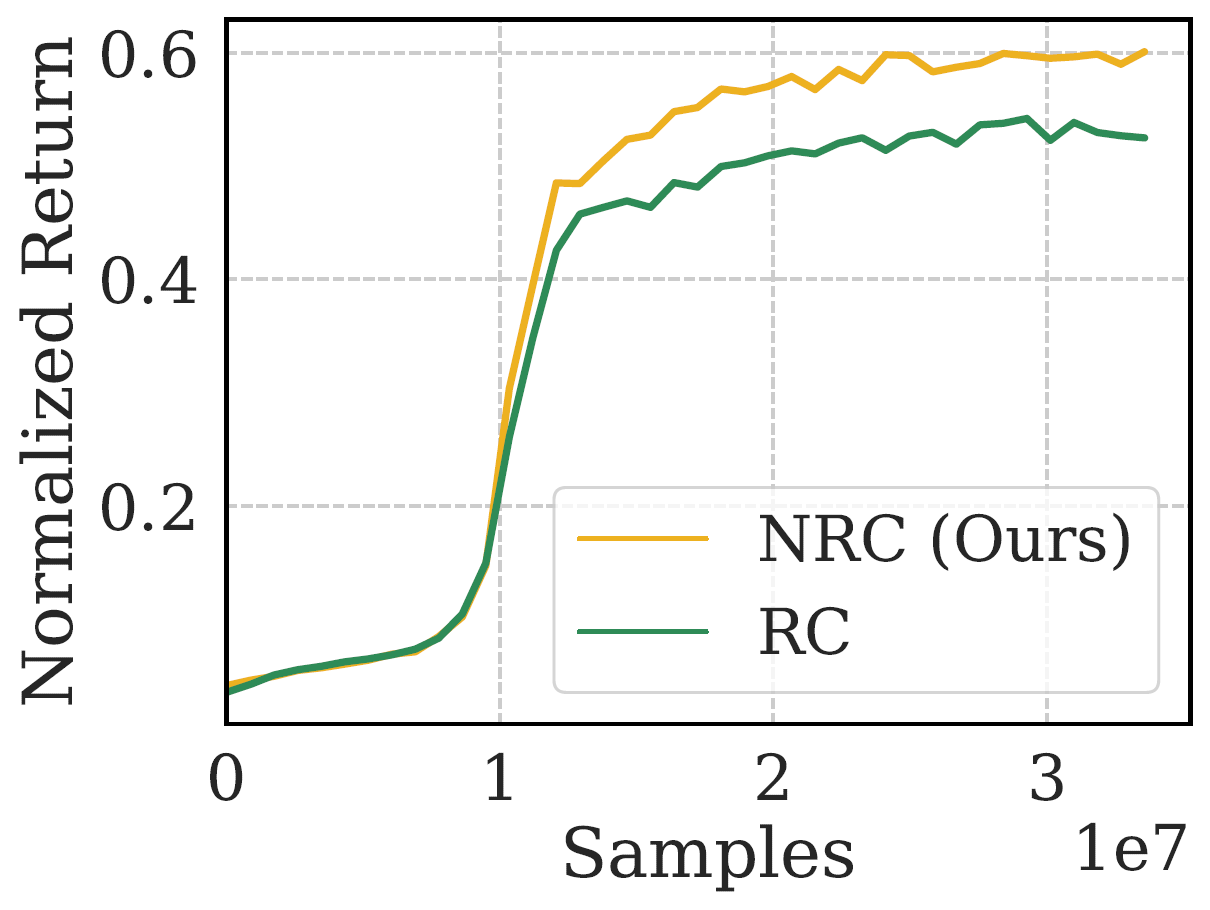}
     \caption{}
     \label{fig:reward_residual}
\end{subfigure}
\caption{Comparison between (a) proposed Curriculum~(CR) and Non-Curriculum~(NCR) methods and (b) proposed Non-Residual Control~(NRC) and Residual Control~(RC) used in previous work~\cite{xie2020learning}. Our proposed method shows best overall training performance in terms of learning speed and converged rewards.
The corresponding total samples for the curriculum is $6.6e7$, while the NCR model has full range of dynamics randomization from the start of training.
}
\label{fig:reward}
\vspace{-0.5cm}
\end{figure}

To evaluate the effects of the randomization curriculum, we compare the performance of policies trained with and without the curriculum. Fig.~\ref{fig:reward_curr} compares learning curves for the different policies. The policy trained with the randomization curriculum (CR) starts training with a small amount of randomization, which is then gradually increased over the course of training. The policy trained without the curriculum (NCR) starts training with the full range of randomization. 
The large amount of randomization at the start of training leads the policy to adopt an excessively conservative and sub-optimal behavior. The policy trained with the curriculum exhibits substantially faster learning progress, while also achieving a higher return. 

The effects of residual control are evaluated by comparing the performance of our policy that uses non-residual control (NRC), with a policy that uses residual control (RC) \cite{xie2018feedback,xie2020learning}. Learning curves comparing the different policies are available in Fig.~\ref{fig:reward_residual}. In the absence of external perturbations, the performance of the two policies is similar, with the non-residual policy performing marginally better than the residual policy. However, as we show in the following experiments, our non-residual policy with dynamics randomization is more robust than the residual policy, which may be due to the non-residual policy's greater flexibility to deviate from the behaviors prescribed by the reference motion in order to recover from perturbations.

\subsection{Robustness Analysis in High-Fidelity Simulation}
A \emph{Feasible command set} is a set of input gait parameters $p^d$ that will not cause a controller to fail.
A \emph{safe set} is defined as a set of gait parameters that a controller actually achieves on the robot while maintaining a stable walking gait.
During each iteration, a gait parameter $p^d$ is provided to the controller as a command in MATLAB SimMechanics,
if the controller succeeds in maintaining a stable gait for 15 seconds, then $p^d$ will be added to the feasible command set and the actual achieved gait parameter $\hat{p}$ will be added to the safe set. 
Through extensive tests of a controller, the resulting feasible command set and safe set provide informative metrics to evaluate the control performance and robustness of the controller when deployed on the robot.
Typically, a walking controller with a larger feasible command set can handle more scenarios, and a controller with a larger safe set can achieve more dynamic motions. 
Moreover, a controller with better tracking performance can result in a similar shape between the feasible command set and safe set, as the difference between these two sets indicates tracking errors between $p^d$ and $\hat{p}$. 

\begin{figure}[!tb]
    \centering
    \begin{subfigure}[b]{0.49\columnwidth}
         \includegraphics[width=\textwidth]{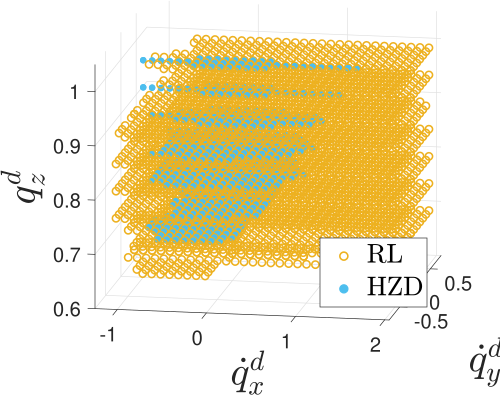} 
         \caption{Feasible Command Set}
         \label{subfig:feasible_commands_set_3d}
    \end{subfigure}
    \begin{subfigure}[b]{0.49\columnwidth}
         \includegraphics[width=\textwidth]{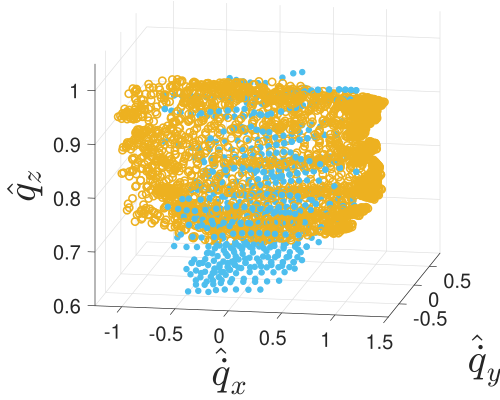}
         \caption{Safe Set}
         \label{subfig:safe_set}
    \end{subfigure}
    \caption{Comparison of proposed commands (Feasible Command Set) and achieved commands (Safe Set) between HZD-based controller~\cite{li2020animated} and proposed RL-based controller. Our RL-based controller can handle more  tracking commands than the HZD-based baseline and thus results in a larger feasible command set. The safe set of RL-based policy is also larger in the sagittal walking velocity $\hat{\dot{q}}_x$ and walking height $\hat{q}_z$ direction. The tracking performance of the RL-based policy also shows advantages as the shapes of feasible command set and safe set are closer.}
    \label{fig:feasible_safe_sets}
    \vspace{-0.5cm}
\end{figure}

We compare the feasible command sets and safe sets between the learned policy and prior HZD-based variable walking height controller developed in~\cite{li2020animated} and based on~\cite{gong2019feedback}. 
The procedures for generating the command sets and safe sets of these two controllers are identical, and the testing range for $p^d$ is set to be between $[-1.1, ~-0.6, ~0.65]^T$ and $[2, ~0.6, ~1.0]^T$, with a resolution of $[0.1, ~0.1, ~0.05]^T$. The resulting feasible command sets and safe sets are shown in Fig.~\ref{fig:feasible_safe_sets}.
As shown in Fig.~\ref{subfig:feasible_commands_set_3d}, the proposed RL-based controller is able to cover almost the entire testing range, while the HZD-based controller can only handle a smaller bowl-shape region. 
Quantitatively, the feasible command set of the RL-based walking controller is more than 4 times larger than HZD-based controller. 
Moreover, as illustrated in Fig.~\ref{subfig:safe_set}, the RL-based walking controller covers a broader safe set than the HZD-based controller. 
In practice, this means that the RL-based controller can achieve faster forward and backward walking~(from $-1.2$~m/s to $1.2$~m/s) than HZD-based one~(below $1$~m/s). 
Although Fig.~\ref{subfig:safe_set} shows that the HZD-based controller can achieve walking gaits with lower heights~($0.6$~m) than the RL-based one~($0.65$~m), the tracking error of the HZD-based controller is not negligible as its minimum feasible
walking height command can only reach $0.7$~m while RL-based one can go to $0.65$~m. 
Therefore, the RL-based controller exhibits better performance on tracking commands. 
Moreover, by inspecting the relationship between Fig.~\ref{subfig:feasible_commands_set_3d} and Fig.~\ref{subfig:safe_set}, we find that the RL-based controller can handle the commands that are outside of the given gait library in Tab.~\ref{tab:gaitlibrary}, \textit{e.g.}, $2$~m/s in the sagittal direction. The resulting actual velocity $\hat{\dot{q}}_x$ is around $1.2$~m/s, which is also outside of the range seen in the gait library. With the standard HZD-based controller, the robot only approaches $1$~m/s when it is being given a $2$~m/s command, due to a large tracking error. 
This shows that the RL-based controller is not strictly tracking the commands, and instead finds a more optimal gait that is close to the commands while maintaining stability. 

\begin{figure*}[!htp]
    \centering
    \begin{subfigure}{.34\linewidth}
    \includegraphics[width=1\linewidth]{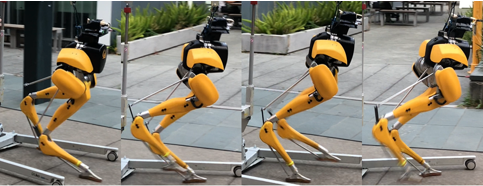}
    \caption{Fast Walking Outdoor}
    \label{subfig:fast_walking}    
    \end{subfigure}
    \begin{subfigure}{.2\linewidth}
    \includegraphics[width=1\linewidth]{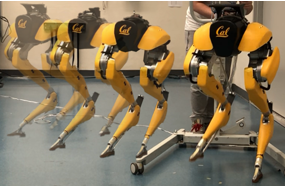}
    \caption{Side Walking}
    \label{subfig:side_walking}    
    \end{subfigure}
    \begin{subfigure}{.44\linewidth}
    \includegraphics[width=1\linewidth]{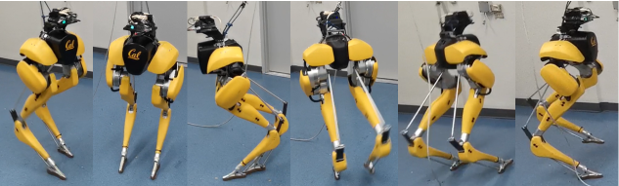}
    \caption{Turning}
    \label{subfig:turning}    
    \end{subfigure}    
    \begin{subfigure}{.53\linewidth}
    \includegraphics[width=1\linewidth]{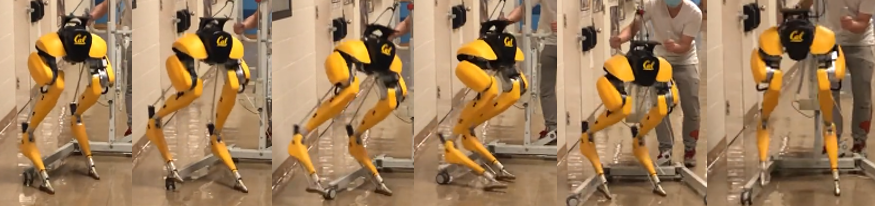}
    \caption{Recover from Foot Sliding}
    \label{subfig:recover_gantry} 
    \end{subfigure}
    \begin{subfigure}{.225\linewidth}
    \includegraphics[width=1\linewidth]{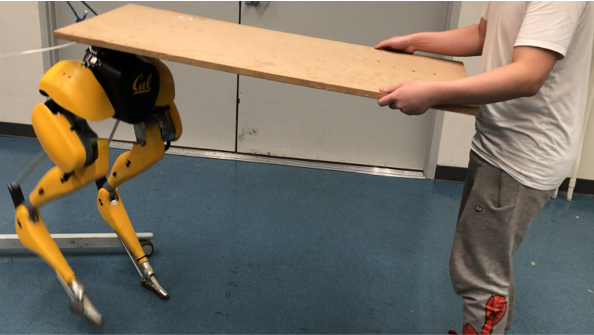}
    \caption{Unknown Load}
    \label{subfig:unknown_load} 
    \end{subfigure}     
    \begin{subfigure}{.112\linewidth}
    \includegraphics[width=1\linewidth]{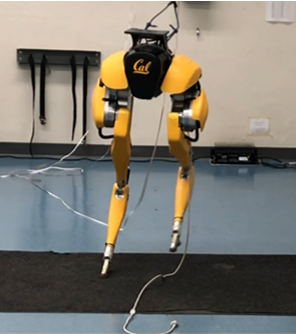}
    \caption{Anti-Slip}
    \label{subfig:anti_slippery_ground} 
    \end{subfigure}
    \begin{subfigure}{.112\linewidth}
    \includegraphics[width=1\linewidth]{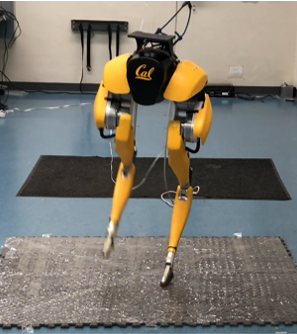}
    \caption{Slippery}
    \label{subfig:slippery_ground} 
    \end{subfigure}        
    \caption{Experiment Results. The proposed learned walking policy extensively on Cassie in real world in different scenarios. In the experiments, the policy enables the Cassie to perform various agile behaviors such as fast forward and backward walking, sideways walking, changing walking height, and turning around. Moreover, empowered by the proposed policy, the robot is able to recover from random perturbation and also able to adapt to change different ground frictions and unknown load.}
    \label{fig:experiments} 
    \vspace{-0.5cm}
\end{figure*}

\subsection{Robustness in the Real World}
The deployed policy on Cassie in the real world can reliably control the robot to perform various behaviors, such as  changing walking heights in~Fig.~\ref{subfig:lower_height},\ref{subfig:enlarge_height}, fast walking in Fig.~\ref{subfig:fast_walking}, walking sideways in~Fig.~\ref{subfig:side_walking}, turning around in~Fig.~\ref{subfig:turning}. 
Moreover, the policy also shows robustness to the changes of the robot itself and the environment.

\subsubsection{Modeling Error}
During the experiments in this paper, a malfunction caused two motors on the Cassie to not work properly. Specifically, the right rotation $q^R_2$ and right knee $q^R_4$ motors were partially damaged, making them unable to produce as much torque as the corresponding motors on the left side or in the simulation.
Following this malfunction, model-based walking controllers, such as the factory default controller and the HZD-based variable walking controller~\cite{li2020animated}, were no longer able to reliably produce a walking gait.
The baseline HZD-based controller was no longer able to recover to a normal height after a reduction in walking height, since the right leg was weaker than the left leg. 
However, by training with dynamics randomization (Sec.~\ref{subsec:domain_randomization}), especially the damping ratio of each joint, the proposed learned walking controller could control the robot even with partially damaged motors. 
Indeed, this policy was able to successfully control the robot the very first time it was deployed, without additional tuning.

\subsubsection{Perturbation}
\begin{figure}[t]
    \centering
    \includegraphics[width=0.6\columnwidth]{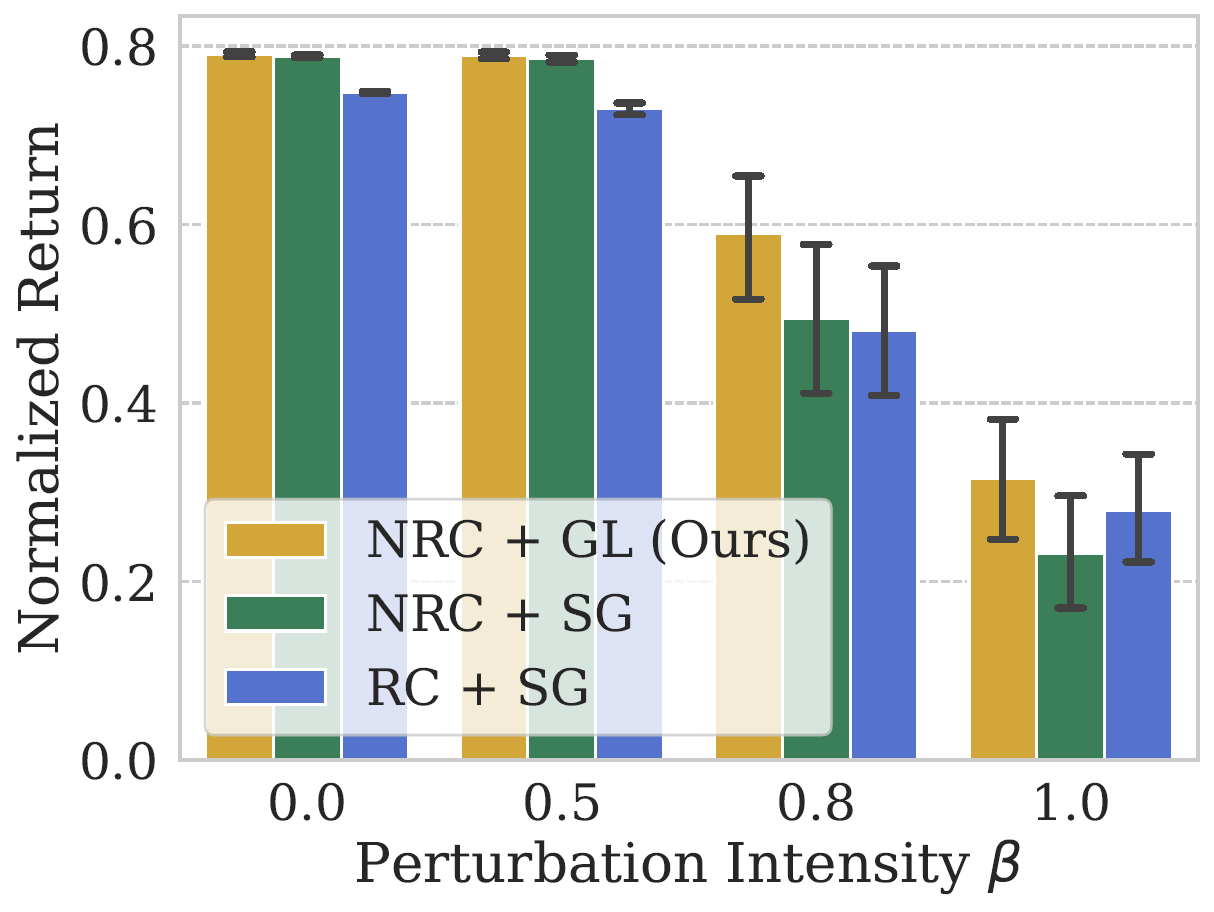}
    \caption{Comparison of robustness to perturbation among 3 different methods in MuJoCo simulation. Non-Residual Controlled Gait Library (NRC+GL) is trained with the gait library; Non-Residual Controlled Single Gait (NRC+SG) and Residual Controlled Single Gait (RC+SG) are trained with only one single gait, which is also the test motion. 
    A $6$ DoF force is randomly applied on the pelvis with probability $0.15\beta\%$. 
    The \textit{Normalized Return} is computed by the mean reward of 32 roll-outs for each model for each $\beta$.}
    \label{fig:comparison}
    \vspace{-0.5cm}
\end{figure}

To show that our approach is more robust, three quantitative experiments are done in the MuJoCo simulator: 1) a non-residual policy trained with the gait library, 2) a non-residual policy trained using a single reference motion from the gait library, and 3) a residual policy trained with the same single gait as the previous work~\cite{xie2020learning}.
All policies are trained without domain randomization.
During the evaluation, the pelvis is perturbed randomly by a $6$ DoF force with a probability of $0.15\%\beta$ at each time step lasting for a random time span sampling from $[0, 0.8\beta]$~s, where $\beta\in[0,1]$ stands for perturbation intensity. The achieved return of each model is illustrated in Fig. \ref{fig:comparison}. When larger perturbations like  $\beta\in\{0.8,1\}$ are applied, the model trained by the proposed method shows significant advantages over other models.

To further demonstrate the robustness in the real world qualitatively, we randomly push Cassie with a rod in different directions, \textit{e.g.} from the front of the pelvis in Fig.~\ref{subfig:recover_front}, from the back in Fig.~\ref{subfig:recover_back}, and from left and right of the pelvis. 
The feet of Cassie are also perturbed during walking, including stepping on the gantry in Fig.~\ref{subfig:recover_gantry}. 
In addition an unknown load is applied in Fig.~\ref{subfig:unknown_load} and changes in ground friction in Fig.~\ref{subfig:anti_slippery_ground},\ref{subfig:slippery_ground}. 
The proposed learned policy shows improved robustness over previous work across all scenarios.

\section{Conclusion and Future Works}
\label{sec:conclusion}
To our knowledge, this paper is the first to develop a diverse and robust bipedal locomotion policy that can walk, turn and squat using parameterized reinforcement learning.
In this work, a model-free reinforcement learning method is proposed to train a policy that is able to control Cassie to track given walking velocities, walking heights, and turning yaw velocities, by imitating reference motions decoded from a HZD-based gait library. 
In contrast to prior work on reinforcement learning for bipedal locomotion with Cassie, our method does not utilize a residual control term, providing improved flexibility, and instead uses the HZD-based gait library to provide references for diverse training. This results in better performance and robustness, as well as more sophisticated recoveries.
The proposed learning method shows benefits over a baseline model-based walking controller, producing a larger feasible command set, a larger safe set, and better tracking performance.
In the real world experiments, the policy also demonstrates considerable robustness, effectively controlling Cassie, even with malfunctioning motors, to walk over floors with different friction and rejecting perturbations.
An exciting future direction is to explore how more dynamic and agile behaviors can be learned for Cassie, building on the approach presented in this work.


\balance
{
\bibliographystyle{IEEEtran}
\bibliography{bib/bibliography}
}


\end{document}